\documentclass[eat,twocolumn]{jmlr}

\graphicspath{{./figures/}} 
\usepackage{mdwmath} 
\usepackage{mdwtab}
\usepackage{eqparbox}
\usepackage{mathtools}
\usepackage[utf8]{inputenc}
\usepackage{dsfont}
\usepackage{bm}
\usepackage{physics}  
\usepackage{csvsimple}  
\usepackage{listings} 
\usepackage{amsfonts}
\usepackage{enumitem}
\setlist[enumerate]{nosep}

\usepackage{gensymb} 
\usepackage{mathtools}

 \usepackage{longtable}

\usepackage{booktabs}

\newcommand{\sqb}[1]{\left[{#1}\right]} 
\newcommand{\nn}{\nonumber} 
\renewcommand{\norm}[1]{\left\lVert{#1}\right\rVert}

\makeatletter
\renewcommand*{\@opargbegintheorem}[3]{\trivlist
  \item[\hskip \labelsep{\bfseries #1\ #2}] \textbf{(#3)}\ \itshape}
\makeatother
\DeclareMathOperator{\E}{\mathbb{E}}

\allowdisplaybreaks

\theorembodyfont{\upshape}
\theoremheaderfont{\scshape}
\theorempostheader{:}

\jmlrvolume{}
\firstpageno{1}

\jmlryear{2022}
\jmlrworkshop{Machine Learning for Health (ML4H) 2022}

\title[CardiacGen]{CardiacGen: A Hierarchical Deep Generative Model for Cardiac
Signals}

\author{\Name{Tushar Agarwal\nametag{\thanks{Corresponding Author}}}
\Email{agarwal.270@osu.edu}\and \Name{Emre Ertin} \Email{ertin.1@osu.edu}\\
\addr Electrical and Computer Engineering, The Ohio State
University, Columbus OH, USA}

\begin{document}

\maketitle


\begin{abstract}
	We present CardiacGen, a Deep Learning framework for generating
	synthetic but physiologically plausible cardiac signals like ECG.
	Based on the physiology of cardiovascular system function, we propose a
	modular hierarchical generative model and impose explicit regularizing
	constraints for training each module using multi-objective loss functions. The model
	comprises 2 modules, an HRV module focused on producing realistic
	Heart-Rate-Variability characteristics and a Morphology module focused on
	generating realistic signal morphologies for different modalities. We empirically show that in addition
	to having realistic physiological features, the synthetic data from CardiacGen
	can be used for data augmentation to improve the performance of Deep
	Learning based classifiers. CardiacGen code is available at
	\url{https://github.com/SENSE-Lab-OSU/cardiac_gen_model}.
\end{abstract}


\section{Introduction}\label{cg.intro}

There are many Machine Learning (ML) algorithms that use physiological signals
like Electrocardiogram (ECG) for useful inference tasks such as biometrics
\citep{rathore2020survey}, Heart-Rate prediction \citep{reiss2019deep} and
stress detection \citep{schmidt2018introducing}. However, the publicly available
datasets are limited in size, especially for training large-scale data-driven
Artificial Neural Networks (ANN) models. Moreover, collecting data at abnormal
physiological conditions such as extreme Heart-Rates and Stress have practical
limitations since most people usually have normal physiological conditions.
Therefore, there is a potential for creating synthetic training data to improve
performance of such Deep Learning (DL) models through data-augmentation.
Alternatively, data-augmentation can be interpreted as an indirect method to
incorporate domain knowledge into the learning process through this synthetic
data. While some parametric models of the data-generation process have been
suggested for cardiac signals
\citep{mcsharry2003dynamical,jafarnia-dabanloo2007modified,zeeman1973differential,
roy2020parameter}, they have limitations especially for augmenting datasets 
as shown by \citet{golany2020simgans}. This is expected as it
is difficult to capture the complexity of interactions in biological dynamical
systems such as our cardiovascular system. 

ANN-based Generative Adversarial Networks (GAN) proposed by
\citet{goodfellow2014generative} are a possible data-driven solution for
data-augmentation, but this immediately raises a question. If the training
dataset isn't large enough to learn a DL classifier, how
will we use it to learn a GAN? Since generative models like GANs try to
approximate the data-generating process, we can better regularize by incorporating domain knowledge
into them, e.g. through appropriate loss functions. Such data-driven models have
received much attention recently
\citep{kuznetsov2020electrocardiogram,zhu2019electrocardiogram,delaney2019synthesis,golany2020simgans}.

Unlike existing DL methods, we use the knowledge about Heart Rate
Variability (HRV) characteristics and pulse morphological properties of cardiac
signals to create a modular generative model called CardiacGen (short for
Cardiac-signal-Generator). CardiacGen comprises two ANN modules based on
Wasserstein GANs with gradient-penalty (WGAN-GP) \citep{gulrajani2017improved}
and its modular design has many advantages described in the following section
\ref{cg.method}. In addition to being used in conventional simulator
applications such as simulated testing of algorithms and privacy preservation,
CardiacGen can be employed to produce cardiac signals under rarely observed
physiological conditions (e.g. very low/high heart rates) and to transcode from
one cardiac modality to another. Moreover, it can enrich existing datasets by
generating samples for underrepresented population groups, potentially reducing
health disparities \citep{martin2019clinical, gao2020deep}. However, we'll focus
on data-augmentation as its primary application in this paper.

Formal definitions of terms Data-Augmentation and Conditional Generative
Modeling, as used in the context of this paper, are provided in appendix Section
\ref{apd:def}. Next, we describe important notation used throughout the text.
The lower-case character $x$ represents a fixed sample value of a random vector
(rv) $X$ (the corresponding upper-case character), $M:N$ is the enumeration of
natural numbers from $M$ to $N$, $g(X_{1:N};w)$ denotes function $g$ of rv's
$X_{1:N}$ with deterministic parameters $w$ and an index-based $x[v]$ represents
the $v^{th}$ value from the ordered set $\{x[i]: i \in \{1:N\}\}$.

Our aim is to design a data-augmenting transformation $A$ as an ANN-based
conditional generative model $CG$ with parameters $w_{CG}$, that generates
random samples of $X$ (a cardiac signal like ECG) conditioned on some $Y$
and a latent rv $Z$, i.e. sampling from $CG(Y,Z;w_{CG})$ must be a good
approximation of sampling from $p(X|Y)$. The dataset of all samples is split
into 3 subsets $\mathcal{D}_{train}$, $\mathcal{D}_{val}$ and
$\mathcal{D}_{test}$ for model training, cross-validation and evaluation
respectively. To augment $\mathcal{D}_{train}$, for every $ y \in
\mathcal{D}_{train}$ we use $CG$ to generate $N$ new samples
$(CG(y)[k],y)_{k=1}^N$ where $CG(y)[k]$ denotes $k^{th}$ stochastic sample of
$CG(y)$.

\section{Method}\label{cg.method}

The human heart consists of 4 chambers, the upper two called atria and lower
two called ventricles. The right two chambers pump 
deoxygenated blood and the left two chambers pump oxygenated blood. For doing
this, the heart goes through a cardiac cycle of contraction
(called systole) and relaxation (called diastole). The timing of contraction
and relaxation events is triggered by electrical signals originating in
specialized cells of the Sino-Atrial (SA) node. Hence, all 4 chambers undergo
corresponding depolarization/repolarization cycles which form the
characteristic morphology comprising peaks and valleys of an ECG signal labeled
as P, Q, R, S, and T (refer appendix Figure \ref{fig:cg.ECG_morph}). The R-peak is the
greatest absolute magnitude point in the signal and is commonly used to locate the cardiac
cycles in the ECG signal. The interval between consecutive R-peaks is called the
RR-interval. Instantaneous Heart Rate (HR) is defined as the inverse of 
RR-interval.

Notably, the timing of the SA node is in turn controlled by sympathetic and
parasympathetic systems of the Autonomic Nervous System. This forms the basis of
inter-beat interval variations known as Heart Rate Variability (HRV). All
low-frequency (LF) variations (0.04-0.15 Hz.) are caused by the sympathetic
system while high-frequency (HF) modulations (0.15-0.4 Hz.) are caused by both
systems, though predominantly by the parasympathetic system. 

Based on this knowledge of the heart's physiology, we modularize the CardiacGen model
$CG$ into 2 ANN based modules, $CG_{HRV}$ and
$CG_{Morph}$. This imposes a hierarchical
structure to the function class in addition to creating a meaningful
intermediate representation. As the
names suggest, $CG_{HRV}$ models HRV characteristics while $CG_{Morph}$ models
morphological properties. It is also important to be able to extract and condense HRV
features from cardiac signal $X$ (ECG) into a uniformly resampled
RR-tachogram representation \citep{electrophysiology1996heart}
$X_{HRV}[i]=R[i]-R[i-1]$ where $R[i]$ denotes the time of $i^{th}$ R-peak
occurrence in seconds. $X_{HRV}$ is further low-pass filtered with a cutoff
frequency of 0.5 Hz. to remove noise. This enables training of the
$CG_{HRV}$ module. There are several advantages of such
modularization including:

\begin{enumerate}
	\item Ability to enforce signal properties via
	penalty terms in the loss function.
	\item May help in privacy preservation applications by enabling masking of HRV or morphological characteristics that might be used to
	identify subjects.
	\item Meaningful intermediate representation of RR-tachogram adds interpretability and can help locate the problematic module in case of a training failure.
	\item Train smaller models (both modules) with fewer parameters compared to a
	large end-to-end model and with projections of the training data tailored to
	appropriate time-scale.
	\item May enable signal imputation and signal transcoding, e.g. obtaining
	Photoplethysmogram (PPG) from ECG using $CG_{Morph}$.
	\item HRV module will facilitate easier knowledge transfer to new datasets         because RR-tachogram representation is robust to most data-set variations.
\end{enumerate}
We use the publicly available WESAD dataset \citep{schmidt2018introducing} for
training CardiacGen. Details of the dataset are described in appendix Section
\ref{cg.data}. The windows of ECG signal being modeled are referred as $X$. Conditioning
vector $Y$ is the concatenation of one-hot representations of available emotion
labels $Y_{emo}$, identity labels $Y_{id}$ and a conditioning signal
$X_{avgHRV}$ or $X_{peak}$ described in the next section.

\begin{figure*}
	\centering
	\subfigure[CardiacGen Model]{\label{fig:cg.model}\includegraphics[width=0.58\textwidth,keepaspectratio]{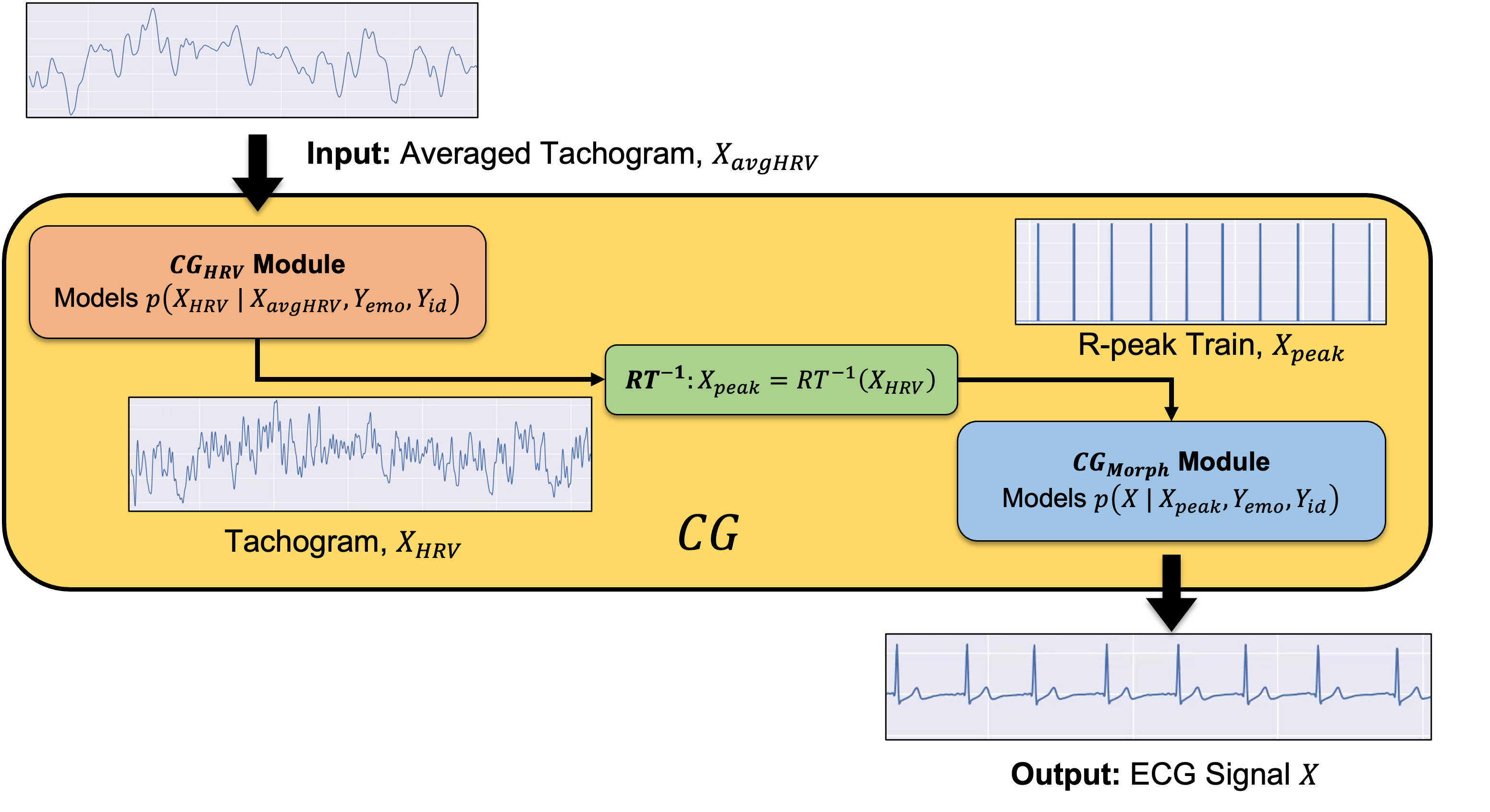}}
	\hfill
	\subfigure[Continuity Aware Training]{\label{fig:cg.continuity}\includegraphics[width=0.41\textwidth,keepaspectratio]{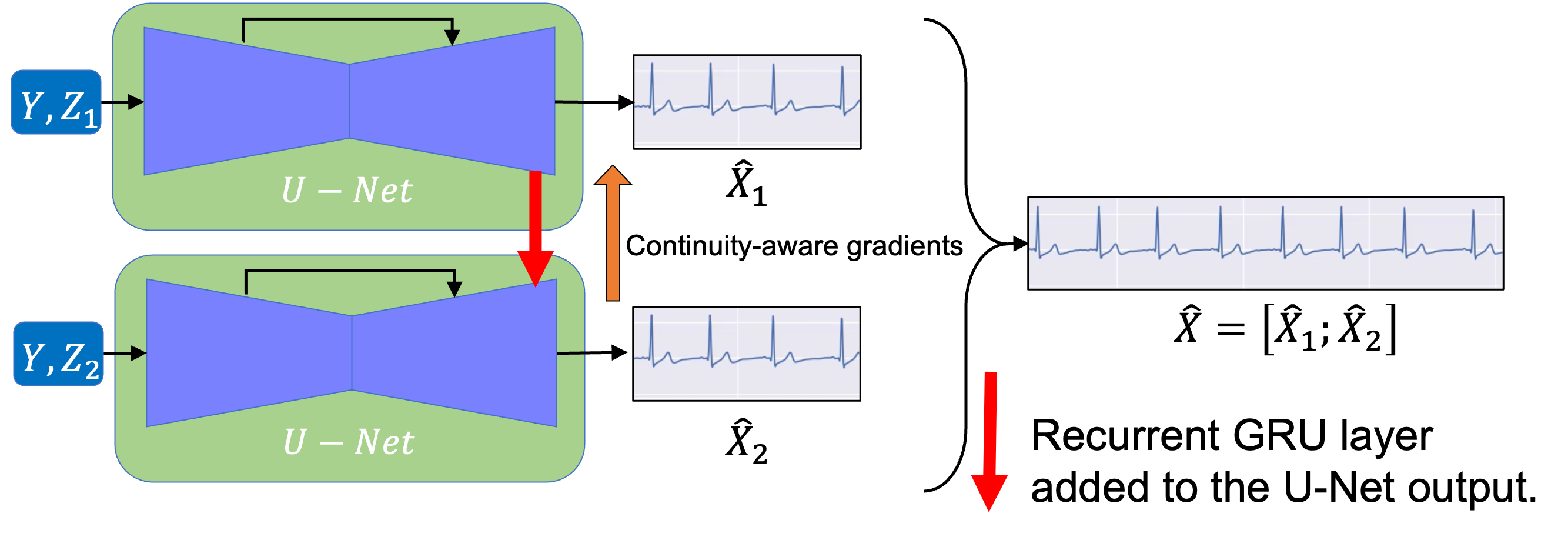}}
	\caption{Schematics for CardiacGen Model and continuity aware training of the generators. }
	\label{fig:results_TAu}
\end{figure*}

\section{Model}\label{cg.model}
CardiacGen consists of two ANNs, the HRV module
$CG_{HRV}(Y_{HRV},Z_{HRV};w_{HRV})$ which creates samples from
$p(X_{HRV}|Y_{HRV})$ and the Morphology module \\
$CG_{Morph}(Y_{Morph},Z_{Morph};w_{Morph})$ which creates samples from
$p(X|Y_{Morph})$ as shown in Figure \ref{fig:cg.model}. Here,
$Y_{HRV}=[X_{avgHRV};Y_{emo};Y_{id}]$, $Y_{Morph}=[X_{peak};Y_{emo};Y_{id}]$,
notation $[u;v]$ represents concatenation of vectors $u$ and $v$. $X_{avgHRV}$
is obtained from the uniformly resampled RR-tachogram and low-pass filtering it
to 0.125 Hz. cutoff frequency and $X_{peak}$ is the R-peak train i.e. a
binary-valued signal with 1's at R-peak locations else 0's. Both $Z_{HRV},
Z_{Morph} \sim \mathcal{N}(0,I)$ as in a conventional GAN.

$X_{avgHRV}$ is chosen as an input condition to $CG_{HRV}$ since it
represents Low-Frequency (LF) properties of $X_{HRV}$ which remain fairly
consistent across subjects. This allows for the desired sampling for
augmentation (See appendix Section \ref{cg.data.postpro}). $X_{peak}$ is
chosen as an input condition to $CG_{Morph}$ because, in our perspective,
it is a more conducive input representation for $CG_{Morph}$. Going from
$X_{peak}$ to the pulsatile ECG $X$ can be viewed as finding a
time-varying non-linear filter. The transformation $RT: X_{HRV}=RT(X_{peak})$ is
fully determined and used at training time to obtain training samples. But at
inference time, the inverse transformation $X_{peak}=RT^{-1}(X_{HRV})$ is
under-determined. So we constrain it by placing the last 1 uniform-randomly in
the last 0.25 seconds (s.) of $X_{peak}$ and using $X_{HRV}$ curve to get the previous
R-peak locations henceforth. 

Both $CG_{HRV}$ and $CG_{Morph}$ modules consist of WGAN-GP \citep{gulrajani2017improved}. WGAN-GP was chosen
because of its advantages over traditional GANs, especially meaningful
loss curves for cross-validation and reliable training. WGAN-GP requires
an extra critic ANN $D(\cdot,\cdot;w_{D})$ to be learned simultaneously to aid
the learning of the distribution approximators (or generators). Both, the critic
and generators are trained alternately using gradient descent steps.
Regularizing loss terms are added to the loss functions used for training generators of both modules. A Power Spectral Density (PSD) reconstruction
loss $\mathcal{L}_{PSD}$ is added to the loss function $\mathcal{L}_{HRV}$ of
$CG_{HRV}$ as shown in appendix eq. \ref{eq:hrv_loss}. Similarly, a mean-squared reconstruction loss $\mathcal{L}_{MSE}$ is added to
loss function $\mathcal{L}_{Morph}$ of $CG_{Morph}$ as shown in appendix eq. \ref{eq:morph_loss}. The WGAN
$\mathcal{L}_{W}$ and gradient-penalty $\mathcal{L}_{GP}$ parts of loss
functions are from eq. 3 of \citet{gulrajani2017improved} and the complete
description of loss functions is provided in appendix Section \ref{apd:exp}. 

Inspired by the seminal work of \citet{isola2017imagetoimage}, the generators of
both modules have a U-Net style architecture comprising convolutional layers.
However, since we want to generate time-series of arbitrary durations, we do 2
modifications. We add a Recurrent Layer having Gated-Recurrent-Unit (GRU) cells
at the output of the U-Net. This can be viewed as non-linear filtering of
features extracted by Convolutional layers and promotes continuity between
outputs of consecutive windows as GRU states are propagated forward. Next, to
further promote this continuity, we train 2 copies of the generator simultaneously
where the second copy uses the GRU state from the first as shown in 
Figure \ref{fig:cg.continuity}.

\section{Experimental Setup}\label{cg.exp}
We use the Tensorflow (2.1) \citep{abadi2016tensorflow} for our implementations.
More details of software and hardware used can be found in appendix section \ref{apd:exp}.
The data pre-processing for training and
post-processing for results in Section \ref{cg.res} are described in appendix
Sections \ref{cg.data.prepro} and \ref{cg.data.postpro} respectively.
Although CardiacGen has multiple use-cases as described in Section
\ref{cg.method}, we'll evaluate its utility by its ability of data-augmentation.
To this end, we implement and train two supervised ANN multi-class classifiers
from recent state-of-the-art works by \citet{sarkar2021selfsupervised} and
\citet{donidalabati2019deepecg}. $g_{emo}$ is a 4-class classifier (neutral,
stress, amusement, and meditation) for emotion recognition based on the former,
and $g_{id}$ is a 15-class classifier (identifying every subject) for identity
recognition based on the latter. Since HRV features are more important for
emotion recognition \citep{hovsepian2015cstress,schmidt2018introducing} while
morphological features are essential for identity recognition, this strategy 
helps evaluate both modules. Additional details of training these
classifiers are provided in the appendix Section \ref{cg.data.util.app}.

\section{Results}\label{cg.res}

\begin{figure*}[ht]
	\floatconts
	  {fig:cg.res}
	  {\caption{Evaluating various aspects of CardiacGen}}
	  {%
		\subfigure[Conditional Generation Ability]{\label{fig:cg.res.cond}%
		  \includegraphics[width=0.43\textwidth,keepaspectratio]{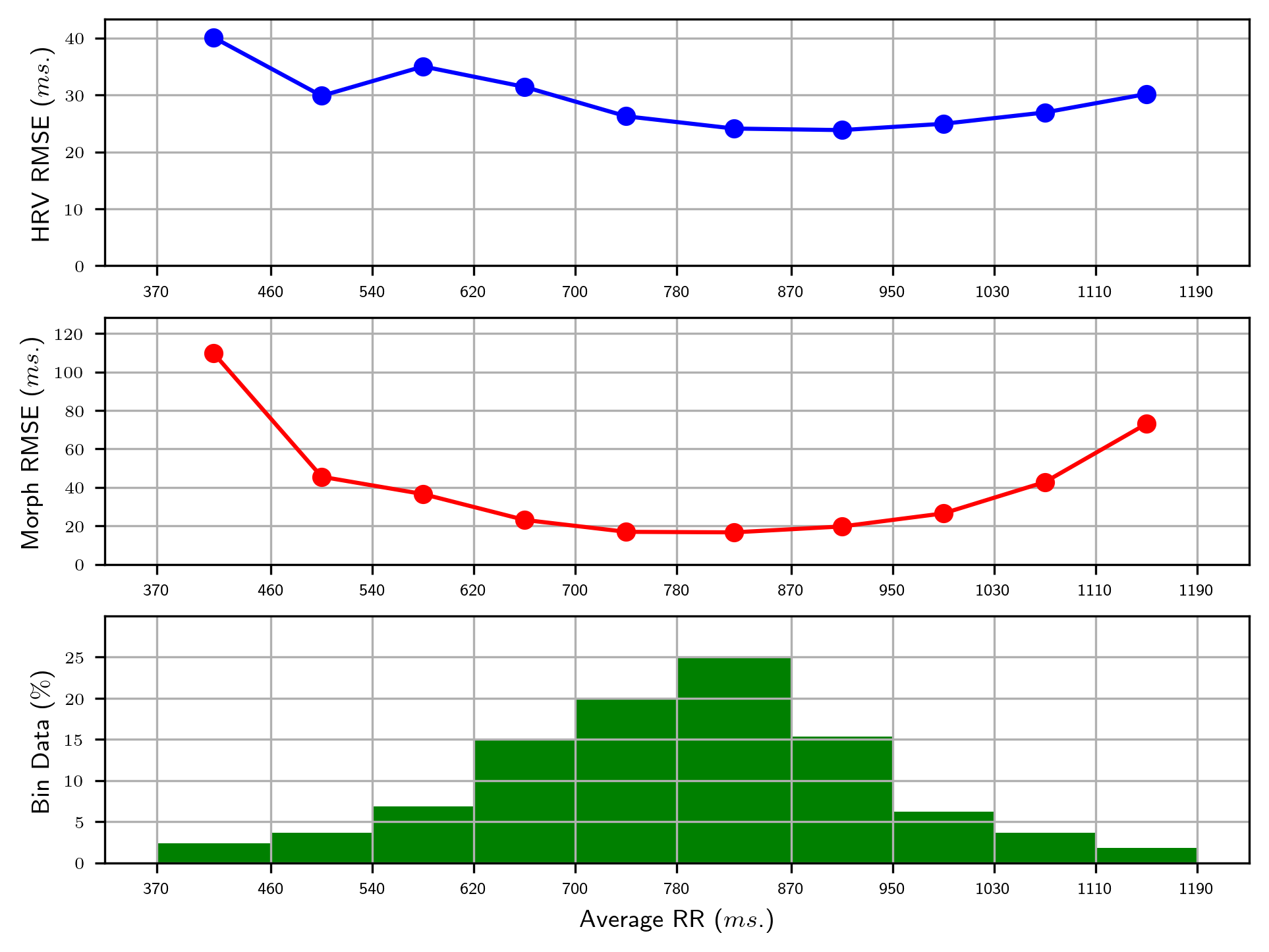}}%
		\quad
		\subfigure[RMSSD for subjects S13 to S17]{\label{fig:cg.res.feat.rmssd}%
		  \includegraphics[width=0.44\textwidth,keepaspectratio]{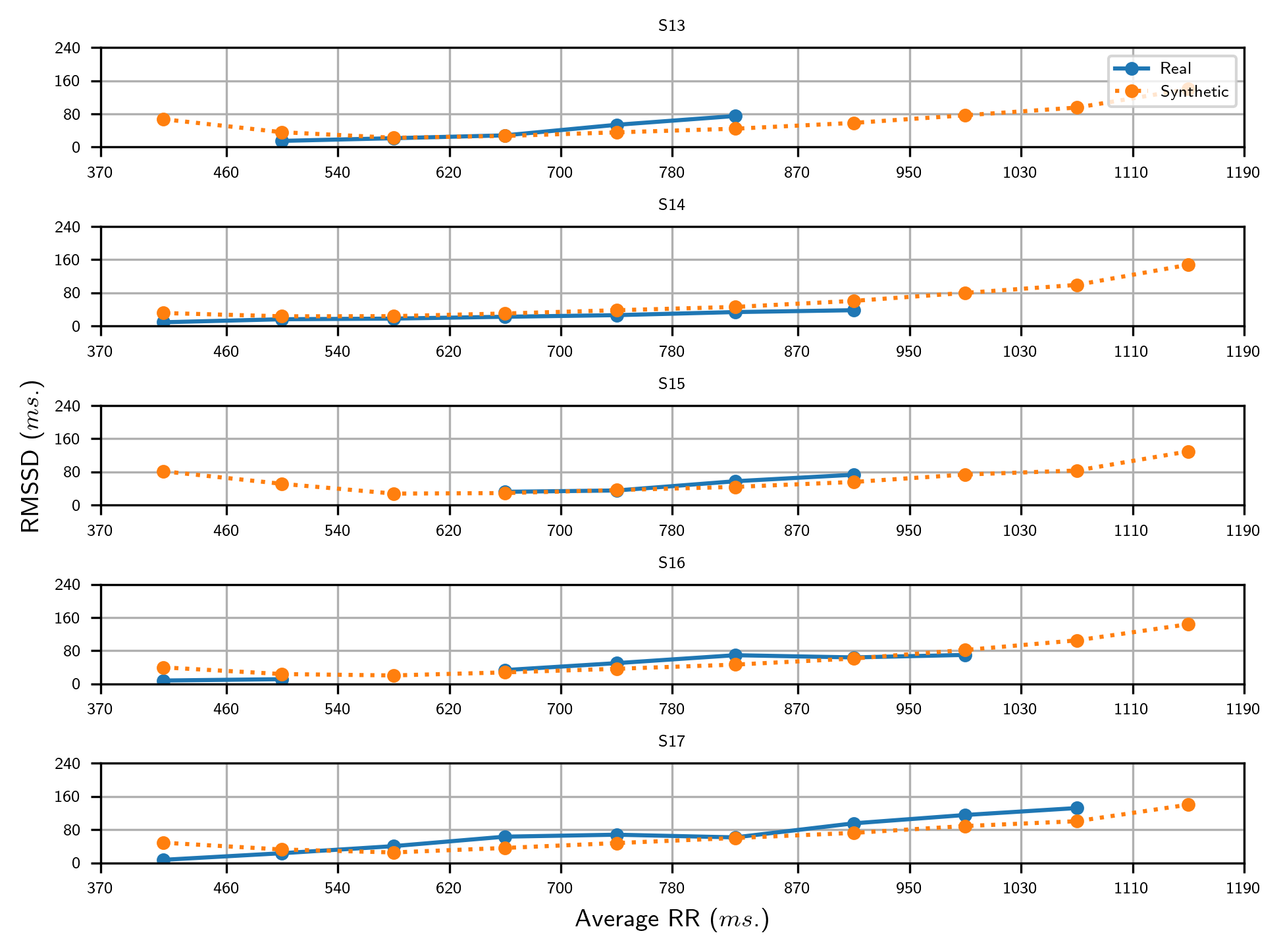}}
		\\
		\subfigure[Test Error for Emotion Recognition]{\label{fig:cg.res.ecg.emo}%
		  \includegraphics[width=0.43\textwidth,keepaspectratio]{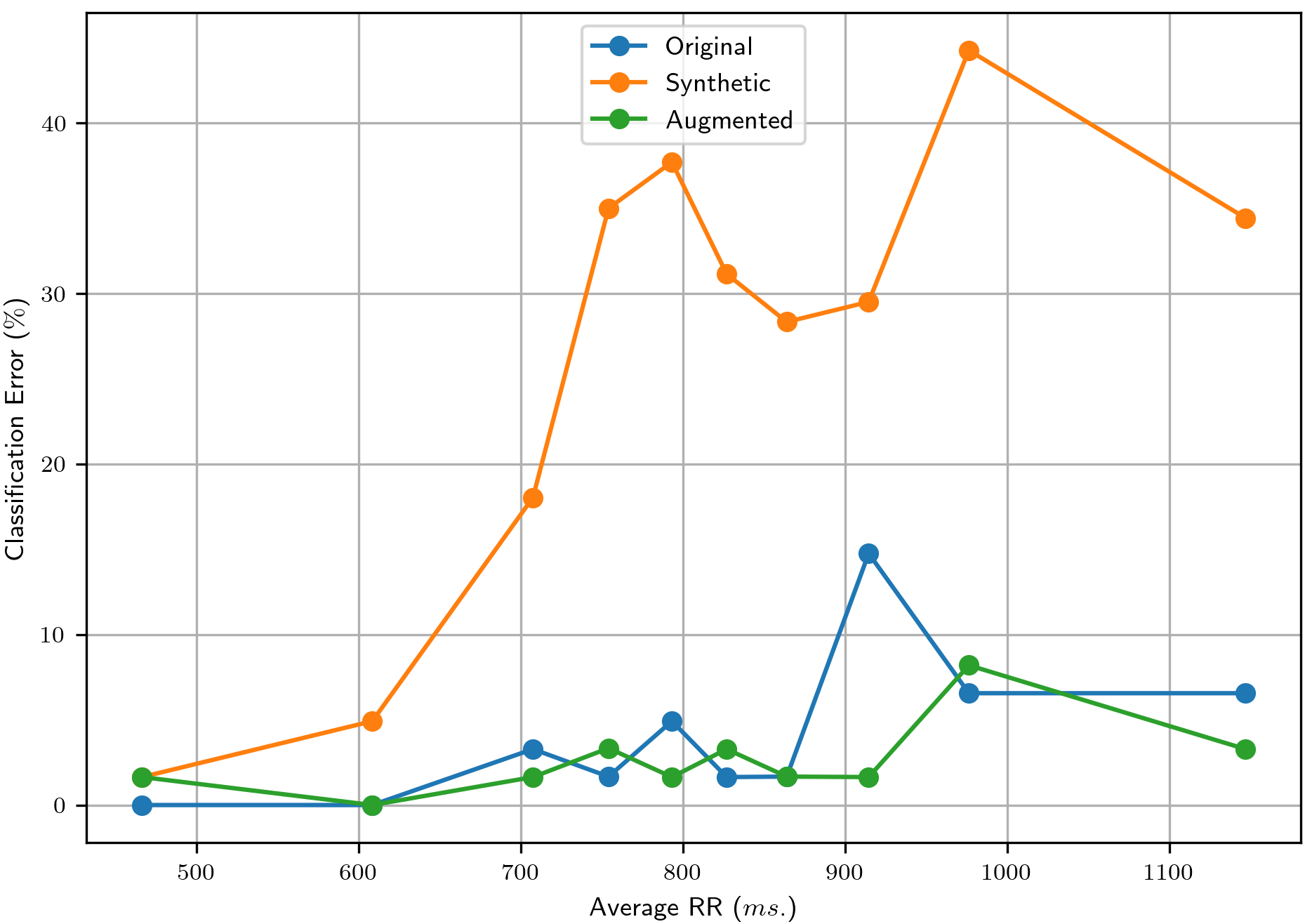}}%
		\quad
		\subfigure[Test Error for Identity Recognition]{\label{fig:cg.res.ecg.id}%
		  \includegraphics[width=0.43\textwidth,keepaspectratio]{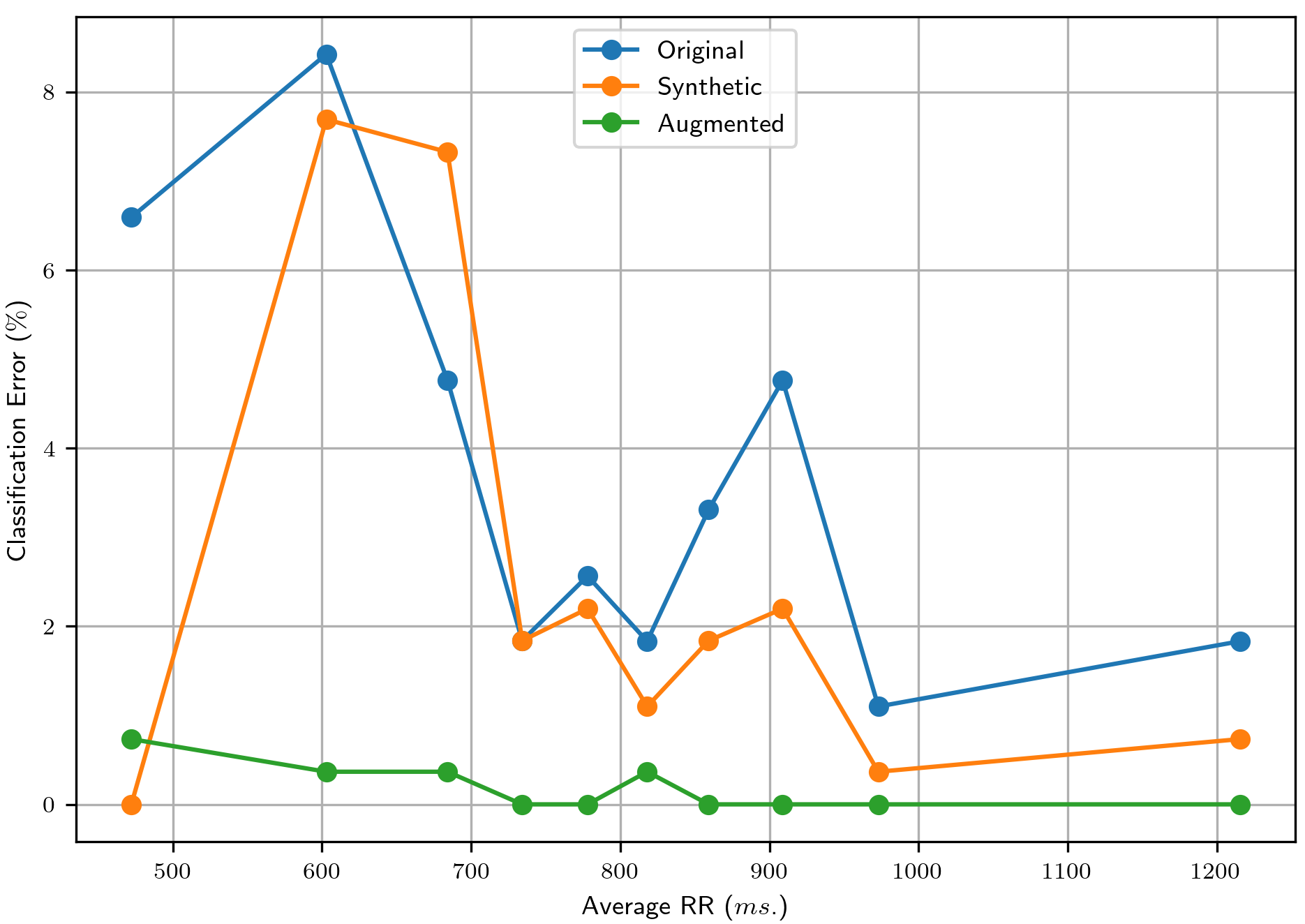}}
	  }
\end{figure*}

In this section, we evaluate CardiacGen based on three quantitative criteria. 
We first test whether the generated output of both CardiacGen modules actually
follows the input real-valued condition. For this, we use the synthetically
generated  $X_{HRV}$ and ECG to re-derive synthetic $X_{avgHRV}$ and $X_{HRV}$
from $CG_{HRV}$ and $CG_{Morph}$ modules respectively. We then find the
root-mean-squared-error (RMSE) between these synthetically re-derived signals
and their real counterparts that were used as input conditions (in ms.). For
Figure \ref{fig:cg.res.cond}, further bin these RMSE values based on the Average
RR length of every window and report the mean RMSE in every bin. It is evident
from Figure \ref{fig:cg.res.cond} that a significant majority of windows have
RMSE lower than 30 ms. for both modules. The overall average RMSE for $CG_{HRV}$
is 27 ms. and for $CG_{Morph}$ is 25 ms. Hence, both modules, and consequently,
CardiacGen, retains information about the real-valued condition vectors well.
The further evaluation of utility will provide a proxy
estimate of conditional generation ability with respect to the categorical
condition vectors.

Next, we verify whether CardiacGen synthesizes realistic physiological data by
calculating HRV feature RMSSD \citep{shaffer2017overview} for 32 s
windows of data and binning them as earlier. Effectiveness of our method is
demonstrated from Figure \ref{fig:cg.res.feat.rmssd} as the empirical
distributions of RMSSD from real and synthetic data are fairly close for all
subjects. The smaller support (i.e. missing bin values) of a subject's
real-data can be explained as follows. A subject
experiences only one sequence of HR during real-data collection while we obtained
synthetic data with HR of other subjects as well, thus increasing its variety.
Results for remaining subjects are provided in appendix section \ref{apd:res}.

The utility of the synthetic ECG from CardiacGen is measured in terms of its
ability to learn ANN models for two classification tasks, emotion and identity
recognition. All the test results use samples from the unseen
$\mathcal{D}_{test}$ after training the classifiers. The values reported are
classification errors, i.e., the percentage of samples misclassified. Graphs
show variations across 10 equally-sized HR bins (i.e. same no. of samples in
each bin).

It is evident from Figure \ref{fig:cg.res.ecg.emo} that the augmented dataset
gives the smallest errors in all bins except one. Even while training entirely
on synthetic data $\mathcal{D}_{synth}$, the errors are much lower than the
random chance error of $75\%$. The overall test errors (in \%) are 4.04, 28.08,
and 2.8 when training using datasets $\mathcal{D}_{train}$,
$\mathcal{D}_{synth}$ and $\mathcal{D}_{aug}$ respectively. Hence, we achieve
more than 30\% improvement after augmentation. Compared with the results of
\citet{sarkar2021selfsupervised}, our classification results establish new
state-of-the-art accuracy for the Affect recognition task using the WESAD
dataset. Although we used the same size of test-set as
\citet{sarkar2021selfsupervised}, we use a fixed test-set instead of the 10-fold
cross-validation done by \citet{sarkar2021selfsupervised} which is a limitation
of our evaluation. This was done to make training feasible for all 
models, in our limited computation and time resources.

Figure \ref{fig:cg.res.ecg.id} further shows the significance of synthetic data
from CardiacGen in improving identity recognition performance. In fact, just
using synthetic data from CardiacGen is at par with using real data in
$\mathcal{D}_{train}$. The overall test errors (in \%) are 1.03, 1.94, 0.18 when
 using datasets $\mathcal{D}_{train}$,  $\mathcal{D}_{synth}$ and
$\mathcal{D}_{aug}$ respectively. Hence, we get more than 82\% improvement after
augmentation.

\section{Conclusion}\label{cg.conc}

We proposed a generative model, CardiacGen, for generating synthetic cardiac
signals. We demonstrated the model's ability to produce physiologically
plausible signals as well as its ability to augment datasets. However, the
effect of the stochastic latent variable $Z$ on CardiacGen's output is
negligible. As a result, the conditional variance of CardiacGen is smaller than
the variance observed in measured signals.  This was observed in pix2pix
\citep{isola2017imagetoimage} as well. Moreover, the WESAD dataset used
for training has a limited size and lacks age as well as gender diversity. In
future work, we plan to improve on these limitations through additional
regularization terms, diverse training datasets as well as extend CardiacGen to generate other cardiac signals like
PPG, and explore signal transcoding
applications e.g. converting ECG to PPG signals.
\acks{This research was partially supported by NSF grants CNS-1823070
CBET-2037398 and NIH Grant P41EB028242}

\bibliography{references}

\appendix

\section{Definitions}\label{apd:def}

\begin{definition}[Data Augmentation]\label{prom:mda}
	Data Augmentation involves applying an appropriate transformation $A:
	\mathcal{D}_{in} \rightarrow \mathcal{D}_{out}$ to a training data-set
	$\mathcal{D}_{train}$ and expand it to an augmented data-set
	$\mathcal{D}_{aug}=A(\mathcal{D}_{train})$. The aim is to find $A$ such that
	the estimated parameters $w$ for a supervised learning problem $\mathcal{P}$ using
	$\mathcal{D}_{aug}$, i.e., $w_{aug}=\mathcal{P}(\mathcal{D}_{aug})$ perform
	better than $w_{train}=\mathcal{P}(\mathcal{D}_{train})$ in terms of the
	chosen real-valued scalar performance measure $\mathcal{M}$.
\end{definition}

\begin{definition}[Conditional Generative Modeling]\label{prom:CGM}
	It aims to estimate the conditional distribution
	$p(X|Y)$  using a parameterized distribution family $q(X|Y;w)$ and a set
	of samples $\mathcal{D}$ from the joint distribution $p(X,Y)$ of random
	vectors $X,Y$. Since $X$ may depend on other factors of variation in
	addition to $Y$, there is inherent stochasticity in the generation of $X$ from
	$Y$. This is typically achieved by sampling a latent random vector $Z$,
	internal to $q(X|Y;w)$, which has a simpler distribution (e.g. standard
	Gaussian).  Therefore, the goal is to generate $N$ realistic samples
	$\{x[v]\}_{v=1}^N$ from $q(X|Y;w)$ as if they were from $p(X|Y)$.
\end{definition}

\begin{figure}[h]
	\floatconts
	  {fig:cg.ECG_morph}
	  {\caption{ECG morphology \citep{agateller2007schematic}}}
	  {\includegraphics[width=0.45\textwidth,keepaspectratio]{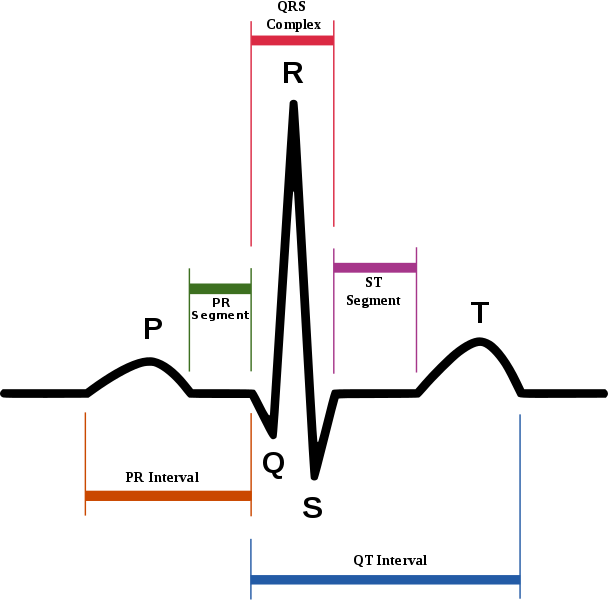}}
\end{figure}

\section{Experimental Details}\label{apd:exp}
The complete loss functions used for training CardiacGen are as follows.

\begin{align}
	\mathcal{L}_{HRV} &= 
	\begin{cases}
		-\mathcal{L}_{W} + \lambda_1\mathcal{L}_{GP}&\text{for } D_{HRV} \\
		\mathcal{L}_{W}  + \lambda_2\mathcal{L}_{PSD}&\text{for } CG_{HRV}
	\end{cases}\label{eq:hrv_loss}
        \\
	\mathcal{L}_{Morph} &= 
	\begin{cases}
		-\mathcal{L}_{W} + \lambda_1\mathcal{L}_{GP}&\text{for }D_{Morph} \\
		\mathcal{L}_{W} + \lambda_3\mathcal{L}_{MSE}&\text{for }CG_{Morph}
	\end{cases}\label{eq:morph_loss}
\end{align}
\begin{align}
	\mathcal{L}_{W} &= \E_{y} \left[ \E_{x}\sqb{D(x,y)}-\E_{\hat{x}}\sqb{D(\hat{x},y)}\right] \nn
	\\
	\mathcal{L}_{GP} &= \E_{(y,\bar{x})}\left[(\norm{\nabla_{\bar{x}}D(\bar{x},y)}_2-1)^2\right] \nn
	\\
	\mathcal{L}_{PSD} &= \E_{y}\left[ \E_{(x,\hat{x})}\left[ \vphantom{\frac{\norm{\abs{\mathcal{F}(x)}^2}_2^2}{N_{\mathcal{F}}}} \right.\right. \nn \\
	& \left.\left. W_{\mathcal{F}} \odot \frac{\norm{\abs{\mathcal{F}(x)}^2-\abs{\mathcal{F}(\hat{x})}^2}_2^2}{N_{\mathcal{F}}}\right]\right] \nn
	\\
	\mathcal{L}_{MSE} &= \E_{y}\left[ \E_{(x,\hat{x})}\left[\norm{x-\hat{x}}_2^2\right]\right] \nn
\end{align}

where $y \sim p(Y)$, $x \sim p(X|y)$, $\hat{x} \sim q(X|y;w)$, $\bar{x} \sim
r(\bar{X}|y)$ is uniform-sampling along straight lines between pairs of sampled
points $x \sim p(X|y), \hat{x} \sim q(X|y;w)$, $\mathcal{F}$ denotes the Fourier
transform operator, $N_{\mathcal{F}}$ is the length of the corresponding Fourier
transform and $W_{\mathcal{F}}$ is a weighting used to emphasize HF components
of the PSD. All expectations are approximated using corresponding empirical
means. Motivated by \citet{gulrajani2017improved}, we employ negative critic
loss as our primary metric for model selection. Hence, the weights of the final
model are set to their values corresponding to the epoch where the smallest
negative critic loss on $\mathcal{D}_{val}$ was achieved. For the
hyperparameters, we use $\lambda_1=10$ (as in \citet{gulrajani2017improved}) and
set $\lambda_2=10^{-2}$, $\lambda_3=5$ to approximately balance the two
regularizing loss terms $L_{PSD}$ and $L_{MSE}$ with the WGAN loss term
$\mathcal{L}_{W}$ in magnitude. Unless mentioned otherwise, the Adam
optimizer from Tensorflow with default parameters is used for most training runs,
including the CardiacGen modules. NeuroKit2 (0.1.0) \citep{makowski2021neurokit2}
python library is used for various physiological signal processing, like R-peak
detection and physiological feature estimation. We use NVIDIA GeForce
RTX 2080 Ti GPU along-with Intel Xeon CPU as our primary computation hardware.

\subsection{WESAD Dataset}\label{cg.data}

Wearable Stress and Affect Dataset (WESAD) contains data from 15 subjects (12
males, 3 females) collected in a laboratory setting. The dataset has several
physiological and motion sensor modalities from both a wrist-worn and a
chest-worn device, but we'll focus on using only ECG signals. We resample it
from 700 Hz to 100 Hz and $X_P$ from 64 Hz to 25 Hz. The dataset also provides 5
emotion labels (0 = undefined/transient/irrelevant, 1 = baseline, 2 = stress, 3
= amusement, 4 = meditation) and 15 subject-ID labels. Note that we aggregated
irrelevant emotion-labels 5/6/7 into 0 as well. We'll use
one-hot-representations $Y_{emo} \in \mathbb{R}^{5}$ of emotion labels and
$Y_{id} \in \mathbb{R}^{15}$ of identity labels. WESAD has signals of approx.
~90 minutes for each subject. The total duration of data used after
pre-processing for ECG is 89,358 s. and the relative proportions of emotion
labels \{0,1,2,3,4\} are \{0.47,0.21,0.12,0.07,0.13\} respectively.

\subsection{Pre-Processing for Training CardiacGen}\label{cg.data.prepro}
While constructing datasets $\mathcal{D}_{train}$, $\mathcal{D}_{val}$ and
$\mathcal{D}_{test}$ for learning as well as evaluating DL models, we create
segments/frames of window-length $win=8 s.$ for ECG signals with a step-size
$step=2 s.$ These segmented input vectors are much more conducive to batch training than arbitrarily
long signal vectors and we refer to the dataset of all such segments as
$\mathcal{D}_{full}$.

While forming $\mathcal{D}_{val}$ and $\mathcal{D}_{test}$, it is challenging to
maintain a variety of all signals. We adopt the following sampling strategy. We
fix block size $bsize=13$ of consecutive segments for $\mathcal{D}_{test}$ and
$\mathcal{D}_{val}$. Then for every subject's $N_{id}$ signal segments, we
sample $N_{bid}$ blocks of $bsize$ consecutive segments uniformly at random such
that $N_{bid} \times bsize\approx 0.1\times N_{id}$ i.e. we select approx. $10
\%$ segments. Let's refer to this sampling process as $UD(N_{id},bsize)$. Hence,
$\mathcal{D}_{test}$ is first constructed with test-segments from such a
sampling. In addition to these test-segments, we drop the 6 overlapping terminal
segments for every block of segments in $\mathcal{D}_{test}$ from
$\mathcal{D}_{full}$ to ensure no data overlap further. From the remaining
segments, we use the same sampling process $UD(N_{id},bsize)$ to get
validation-segments and form $\mathcal{D}_{val}$. We only drop these selected
validation segments from $\mathcal{D}_{full}$ and the remaining segments form
the train segments of $\mathcal{D}_{train}$. Even though there is some overlap
between validation and train segments, the majority of validation segments are
non-overlapping which will ensure meaningful cross-validation results.

\subsection{Post-Processing for Synthetic Data Generation}\label{cg.data.postpro}

We describe our protocol to generate and use synthetic data in this section.  We
produce synthetic ECG data using the joint input condition
$Y_{HRV}=[X_{avgHRV};Y_{emo};Y_{id}]$. $Y_{emo}$ and $Y_{id}$ are one-hot
representations of class numbers sampled from the fixed sets $\{0:14\}$ and
$\{1:4\}$ respectively. Although we use class 0 of $Y_{emo}$ during training, we
don't use it further as it comprises all irrelevant emotion labels. For
$X_{avgHRV}$, $N_{CG}$ continuous windows $\{x_{avgHRV}[v]\}_{v=1}^{N_{CG}}$ can
be sampled from the fixed set
$HV_{\mathcal{D}}=\{x_{avgHRV}[u]\}_{u=1}^{N_{\mathcal{D}}}$ of all windows from
corresponding real dataset $\mathcal{D}$. Thus, for each subject's continuous
windows of $X_{avgHRV}$, we produce synthetic data $\mathcal{D}_{synth}$ by
using all $60$ possible permutations of $Y_{emo}$ and $Y_{id}$, i.e.
$\mathcal{D}_{synth}$ is $60$ times the size of the corresponding real dataset
$\mathcal{D}$.

The overall sampling process for generating synthetic ECG is then 
\begin{align}
	x_{peak}&= RT^{-1} \circ CG_{HRV}(y_{HRV},z_{HRV})\label{eq:hrv_ecg}\\
	x_E&= CG_{Morph}([x_{peak};y_{emo};y_{id}],z_{Morph}\label{eq:morph_ecg})
\end{align}

For comparing the Physiological Features of real data with synthetic data,
$\mathcal{D}_{full}^{32,0}$ consisting of all non-overlapping 32 s. windows of
ECG is used as the Real Dataset and
$HV_{\mathcal{D}_{full}^{32,0}}=\{x_{avgHRV}[u]\}_{u=1}^{N_{\mathcal{D}_{full}^{32,0}}}$
is used to produce corresponding synthetic dataset $\mathcal{D}_{synth}$. The
reason for windowing as such is that HRV features of our interest are usually
evaluated for windows of more than 30 s. in length \citep{kreibig2010autonomic}.
Since our sampling process of $\mathcal{D}_{synth}$  equally distributes all
stress conditions 1 to 4 for every subject, $\mathcal{D}_{synth}$ is
subject-wise sub-sampled such that the ratio of windows with stress conditions 1
to 4 is the same as in $\mathcal{D}_{full}^{32,0}$ of that subject. This is done
so that HRV statistics used for further evaluation are comparable. The results
are described in Section \ref{cg.res}. Errors in the ECG R-peak detection
algorithm cause several artificially high and low RR values beyond
physiologically possible values. These outliers were present in both
$\mathcal{D}_{train}$ and $\mathcal{D}_{synth}$. Removal of 0.5\% smallest and
0.5\% largest RR values help us to avoid these outliers for our evaluations. We
note that the smaller $\mathcal{D}_{test}$ used for utility evaluation didn't
have any such outliers.

For comparing the utility of real ECG data and synthetic ECG data,
$\mathcal{D}_{train}$ described above is used as the Real ECG training Dataset
and $HV_{\mathcal{D}_{train}}=\{x_{avgHRV}[u]\}_{u=1}^{N_{\mathcal{D}_{train}}}$
is used to produce corresponding synthetic ECG training data
$\mathcal{D}_{synth}$. Finally, we form the augmented training dataset
$\mathcal{D}_{aug}=\mathcal{D}_{synth}\cup \mathcal{D}_{train}$ where both
$\mathcal{D}_{synth}$ and $\mathcal{D}_{train}$ contain approximately same no.
of samples obtained by simply repeating $\mathcal{D}_{train}$ samples
appropriately. The same $\mathcal{D}_{val}$ is used for evaluation during all
utility model-trainings and the same $\mathcal{D}_{test}$ is used for their
evaluation after this utility-training. These post utility-training results are
described in Section \ref{cg.res}.

\subsection{Training Deep Classifiers for assessing CardiacGen's Utility}\label{cg.data.util.app}
For ECG based emotion recognition classifier, data-preprocessing and ANN
architecture are taken from \citet{sarkar2021selfsupervised} with minor
modifications for compatibility. ECG is over-sampled from 100 Hz to 256 Hz, i.e.
from the sampling-rate we use to what the authors use. As suggested by the
authors, we segment data into 10 s. windows.

For ECG based identity recognition classifier, data-preprocessing is kept
almost the same as above and only the ANN architecture is taken from
\citet{donidalabati2019deepecg} with some modifications. A stride of 2 is added
to all convolutional layers and Local Response Normalization (LRN) layers are
replaced with Batch Normalization (BN) layers.  We don't derive feature vectors
as in \citet{donidalabati2019deepecg} and directly use signals themselves as
inputs. We use signal windows of size 4 s. Since windows of size 10 s. have
been shown to be sufficient for extracting stress indicating HRV features by
\citet{sarkar2021selfsupervised}, the smaller size windows are used to make
classifier focus on localized morphological properties for this
closed-set-identification problem. Also, the identity recognition
classifier are trained using Stochastic Gradient Descent optimizer with a
learning rate of 0.001 and Nesterov-momentum of 0.6. All classifiers are trained
using the standard categorical cross-entropy loss.

\newpage

\section{Additional Results}\label{apd:res}

\begin{figure}[h]
	\floatconts
	  {fig:cg.res.feat.rmssd.app}
	  {\caption{Evaluating HRV feature RMSSD}}
	  {%
		\subfigure[Subjects S2 to S6]{\label{fig:cg.res.feat.rmssd0}%
		  \includegraphics[width=0.45\textwidth,keepaspectratio]{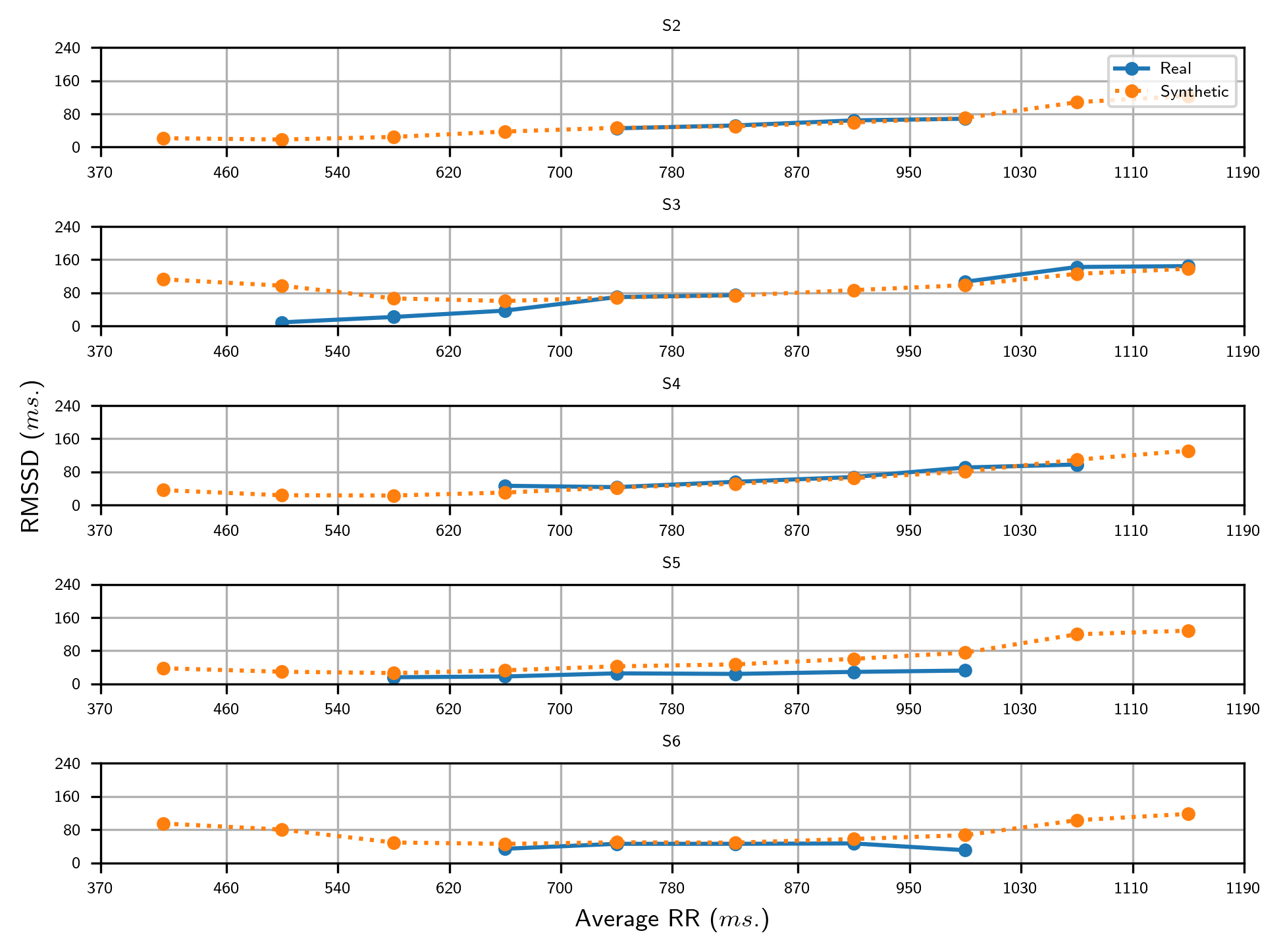}}
		\\
		\subfigure[Subjects S7 to S11]{\label{fig:cg.res.feat.rmssd1}%
		  \includegraphics[width=0.45\textwidth,keepaspectratio]{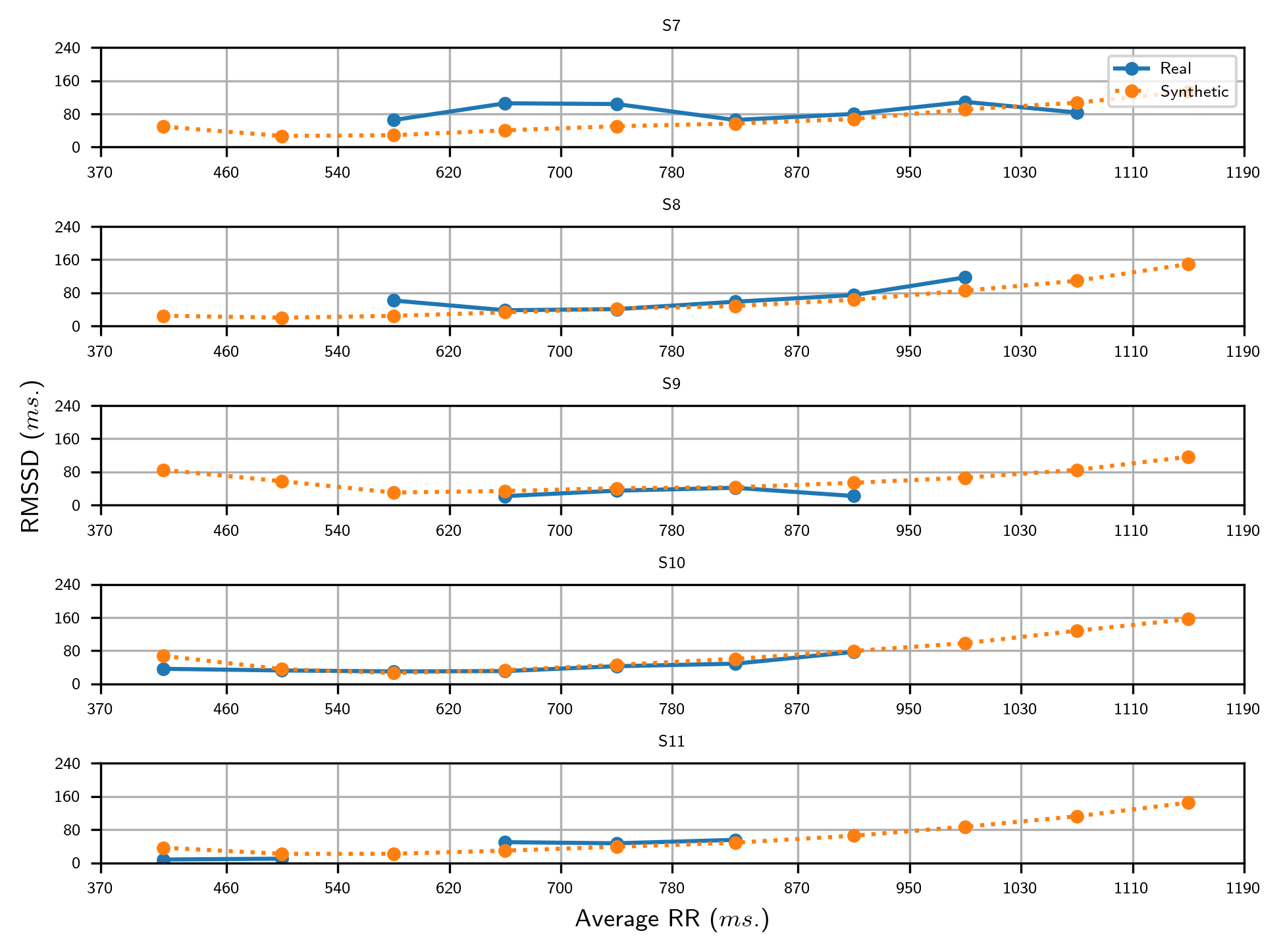}}
	  }
\end{figure}

\end{document}